\documentclass[10pt,twocolumn,letterpaper]{article}

\usepackage{cvpr}              

\usepackage{amsmath,amsfonts,bm}

\def\eqref#1{equation~\ref{#1}}

\def\1{\bm{1}}

\def\vv{{\bm{v}}}

\DeclareMathAlphabet{\mathsfit}{\encodingdefault}{\sfdefault}{m}{sl}
\SetMathAlphabet{\mathsfit}{bold}{\encodingdefault}{\sfdefault}{bx}{n}

\newcommand{\R}{\mathbb{R}}

\usepackage{graphicx}
\usepackage{amsmath}
\usepackage{amssymb}
\usepackage{booktabs}

\usepackage[pagebackref,breaklinks,colorlinks]{hyperref}

\usepackage[capitalize]{cleveref}
\crefname{section}{Sec.}{Secs.}
\Crefname{section}{Section}{Sections}
\Crefname{table}{Table}{Tables}
\crefname{table}{Tab.}{Tabs.}

\usepackage{xcolor}         
\usepackage{makecell}
\usepackage{booktabs}
\usepackage{multirow}
\usepackage{multicol}
\usepackage{mathtools}
\newcommand{\thetav}{\boldsymbol{\theta}}
\newcommand{\phiv}{\boldsymbol{\phi}}

\newcommand{\tv}[0]{{\boldsymbol{t}}}

\newcommand{\Xcal}{\mathcal{X}}
\newcommand{\Ycal}{\mathcal{Y}}

\newcommand{\Bcal}{\mathcal{B}}

\newcommand{\Tcal}{\mathcal{T}}

\newcommand{\Lcal}{\mathcal{L}}

\newcommand{\Pcal}{\mathcal{P}}

\newcommand{\Scal}{\mathcal{S}}

\newcommand{\uv}{\boldsymbol{u}}

\newcommand{\xv}{\boldsymbol{x}}

\usepackage{caption}
\usepackage{subcaption}
\usepackage{enumitem}

\newcommand{\citep}{\cite}

\def\cvprPaperID{12064} 
\def\confName{CVPR}
\def\confYear{2023}

\begin{document}

\title{Prefix Conditioning Unifies Language and Label Supervision}

\author{%
  Kuniaki Saito$^{1,2}$\thanks{Work done during internship at Google Cloud AI Research.}\:,\; Kihyuk Sohn$^{3}$\:,\; Xiang Zhang$^{2}$\:,\; Chun-Liang Li$^{2}$\:,\; \\
  {Chen-Yu Lee$^{2}$\:,\;Kate Saenko$^{1,4}$\:,\;Tomas Pfister$^{2}$}\\
  \texttt{\{keisaito, saenko\}@bu.edu} \\
  \texttt{\{kihyuks,fancyzhx,chunliang,chenyulee,tpfister\}@google.com}\\[.2cm]
  $^{1}$Boston University, $^{2}$Google Cloud AI Research, $^{3}$Google Research, $^{4}$MIT-IBM Watson AI Lab\\
}

\maketitle

\begin{abstract}
Pretraining visual models on web-scale image-caption datasets has recently emerged as a powerful  alternative to traditional pretraining on image classification data. Image-caption datasets are more ``open-domain'', containing broader scene types and vocabulary words, and result in models that have strong performance in few- and zero-shot recognition tasks. However large-scale classification datasets can provide fine-grained categories with a balanced label distribution.
In this work, we study a pretraining strategy that uses both classification and caption datasets to unite their complementary benefits. First, we show that naively unifying the datasets results in sub-optimal performance in downstream zero-shot recognition tasks, as the model is affected by dataset bias: the coverage of image domains and vocabulary words is different in each dataset.
We address this problem with novel Prefix Conditioning, a simple yet effective method that helps disentangle dataset biases from visual concepts. This is done by introducing prefix tokens that inform the language encoder of the input data type (e.g., classification vs caption) at training time. Our approach allows the language encoder to learn from both datasets while also tailoring feature extraction to each dataset. Prefix conditioning is generic and can be easily integrated into existing VL pretraining objectives, such as CLIP or UniCL. In experiments, we show that it improves zero-shot image recognition and robustness to image-level distribution shift.

\end{abstract}
\section{Introduction}
\vspace{-1mm}
\label{sec:intro}
Supervised classification datasets (e.g., ImageNet ~\citep{deng2009imagenet}) have traditionally been used to pretrain image representations for use in downstream tasks. However, web-scale image-caption datasets have recently emerged as a powerful pretraining alternative \citep{radford2021learning,jia2021scaling,li2021align}. Such datasets are more “open-domain", containing a wider variety of scene types and vocabularies than traditional classification datasets, which are biased towards specific categories in their fixed label sets. Consequently, models trained on web-scale image-caption datasets have shown stronger generalization in novel tasks~\citep{radford2021learning, chan2022data} and demonstrated remarkable performance on few and zero-shot image classification tasks~\citep{radford2021learning}.
\begin{figure}
    \centering
    \includegraphics[width=0.45\textwidth]{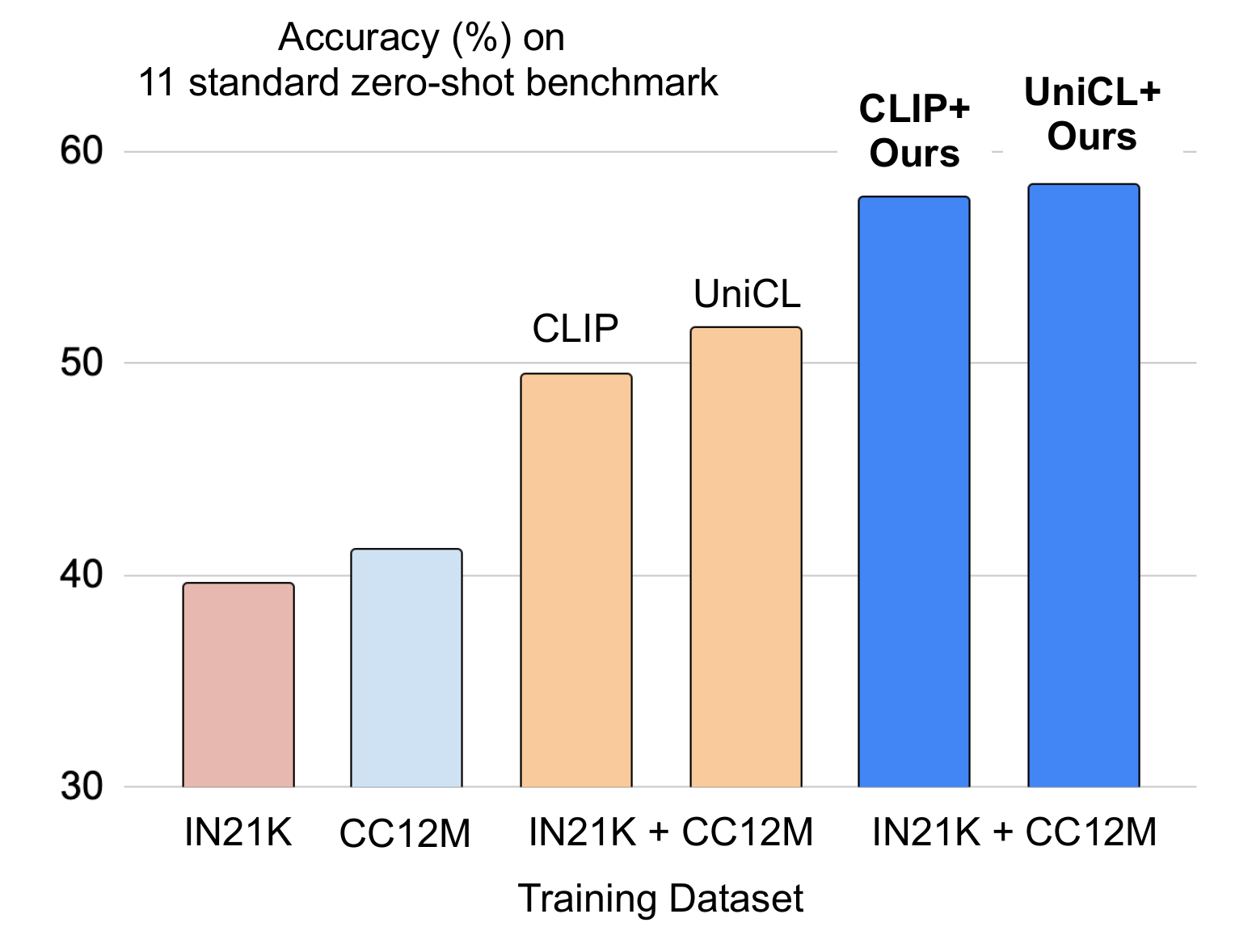}
    \vspace{-3mm}
    \caption{We propose \textit{Prefix Conditioning} to unify image-caption (e.g., CC12M~\citep{changpinyo2021cc12m}) and image classification datasets (e.g., ImageNet21K (IN21K)~\citep{deng2009imagenet}) for training better zero-shot models. Prefix conditioning improves zero-shot recognition performance by more than 6\% on average when training on ImageNet21K and CC12M.}
    \label{fig:teaser}
\end{figure}
Nevertheless, classification datasets are still useful for pre-training as they have a more balanced coverage of categories, including rare and fine-grained categories, and a better focus on the labeled objects in each image. 

Recent works \citep{yang2022unified, yu2022coca} therefore propose to combine caption and classification datasets for pre-training. \cite{yang2022unified} convert classification labels to “label-prompts" by inserting the label into a template sentence, e.g., “a photo of a $<$label$>$."\footnote{We use the term \textit{prompt} to indicate a template sentence filled with a class name.}
Although training on the caption and label-prompt data achieves promising results, it does not fully resolve distribution differences between the open-domain caption data and the classification data. In particular, it produces a language embedding entangled with the classification dataset ``bias''.
We note that classification datasets tend to be biased in at least two ways: 1) the images mostly contain single objects from restricted domains, and 2) the vocabulary is limited and lacks the linguistic flexibility required for zero-shot learning. 
 Therefore, the class embedding of ``a photo of a dog'' optimized for ImageNet may really mean \textit{a photo of a dog from ImageNet} instead, which is biased to ImageNet and does not generalize well to other datasets. 
 We empirically show that such dataset biases negatively affect unified pretraining by reducing the generalization of learned representations and thus jeopardizing zero-shot performance.

To recognize diverse concepts in the open domain, the language model needs to disentangle the dataset bias from the visual concepts and extract language embeddings generalizable to the open domain, e.g., the language embedding representing \textit{a photo of a dog from an open-domain dataset, such as image-caption dataset}, instead of \textit{a photo of a dog from ImageNet}.
Given this intuition, we propose to learn dataset-specific language embeddings, while sharing knowledge from both datasets during training.
We achieve this by a simple yet effective approach we call \textit{Prefix Conditioning}. The idea is to learn a dataset-specific text token (\textit{prefix}) for each dataset so that the bias of the dataset can be absorbed into this token, and in return the remaining text tokens can focus on learning visual concepts. Specifically, we prepend a different token for each dataset (e.g., image classification or caption dataset) to the text input token sequence during pre-training. 

The idea is in part inspired by prefix or prompt tuning~\citep{li2021prefix,lester2021power,zhou2021learning}, which showed  that learnable tokens prepended to the input token sequences of the pre-trained language models are able to learn task-specific knowledge and thus can be used to solve downstream tasks by combining the knowledge of pre-trained large language models and task-specific prefix tokens. 
Our approach differs from prompt tuning in two ways: 1) the proposed prefix conditioning is designed to unify image-caption and classification datasets by disentangling the dataset bias, which is a unique distinction to prompt-tuning works, 2) our approach is applied for VL \textit{pre-training} while the standard prompt tuning is used in fine-tuning. 

In experiments, the proposed simple technique achieves superior performance on zero-shot evaluation if we use the prefix of the caption dataset to get the language embedding at test time as shown in Fig.~\ref{fig:teaser}. 
Meanwhile, inserting the prefix of the classification dataset leads to better performance on classification data.
We also observe a drastic performance improvement when combining our prefix conditioning with the UniCL~\citep{yang2022unified} objective because of their complementarity.
Our contributions are summarized as follows:
\begin{itemize}[leftmargin=*]
    \item We propose novel Prefix Conditioning  \textit{at pre-training time}  to unify image-label and image-caption supervision. It is the first mechanism to use prefixes to condition the source of the dataset during vision language contrastive pre-training, rather than post pre-training.
   \item This simple approach improves zero-shot recognition performance by more than 6\% on average in experiments on ImageNet21K~\citep{deng2009imagenet} and CC12M~\citep{changpinyo2021cc12m}.
   \item Our comprehensive ablation study shows that prefix conditioning enables the model to switch its approach to extracting language features, e.g., attend to different words.
   \end{itemize}
\vspace{-3mm}
\section{Related Work}
\vspace{-1mm}
\label{sec:related}
\textbf{Vision-Language Contrastive Learning.}
Zero-shot recognition is conventionally solved by learning the relationship between visual representations and word embeddings of the class names~\citep{frome2013devise,akata2015evaluation, xian2017zero,xian2016latent,wang2018zero,mensink2014costa,jayaraman2014zero}. 
Vision-language contrastive learning models, such as CLIP~\citep{radford2021learning}, pre-train a model with a large-scale image-caption data (400M) and achieve a remarkable improvement in zero-shot recognition. 
ALIGN~\citep{jia2021scaling} demonstrated the effect of scaling up the size of image-caption data. 
Various techniques have been proposed to improve the data efficiency given a relatively small amount of image-caption data (order of 10M).
ALBEF~\citep{li2021align} employs model distillation and masked language modeling. DeCLIP~\citep{li2021supervision}, SLIP~\citep{mu2021slip} and TCL~\citep{yang2022vision} harness self-supervised contrastive learning. FILIP~\citep{yao2021filip} uses token-to-token contrastive learning rather than the global contrastive learning used in CLIP. BLIP~\citep{li2022blip} generates pseudo captions to diversify the language modality for each image.
Unlike these works that handle only caption-style supervision, we focus on the use of label supervision in vision-language pre-training. Our approach brings orthogonal improvement to the aforementioned works as they seek to improve training on image-caption data. 

UniCL~\citep{yang2022unified} and K-Lite~\citep{shen2022k} unite the image-caption and image-label supervision by converting labels into text with pre-defined template sentences. UniCL leverages a supervised contrastive loss~\citep{khosla2020supervised} for image-label pairs. K-Lite~\citep{shen2022k} utilizes external knowledge from WordNet~\citep{miller1995wordnet} and Wikitionary~\citep{meyer2012wiktionary}. The input noun is augmented with the class hierarchy and definition to enrich the supervision. Our method is complementary to these approaches since both UniCL and K-Lite do not consider the domain shift between datasets. In experiments, we observe a significant performance boost when UniCL is combined with the prefix conditioning.



\textbf{Learning with Prompts.}
\begin{figure*}
    \centering
    \includegraphics[width=\textwidth]{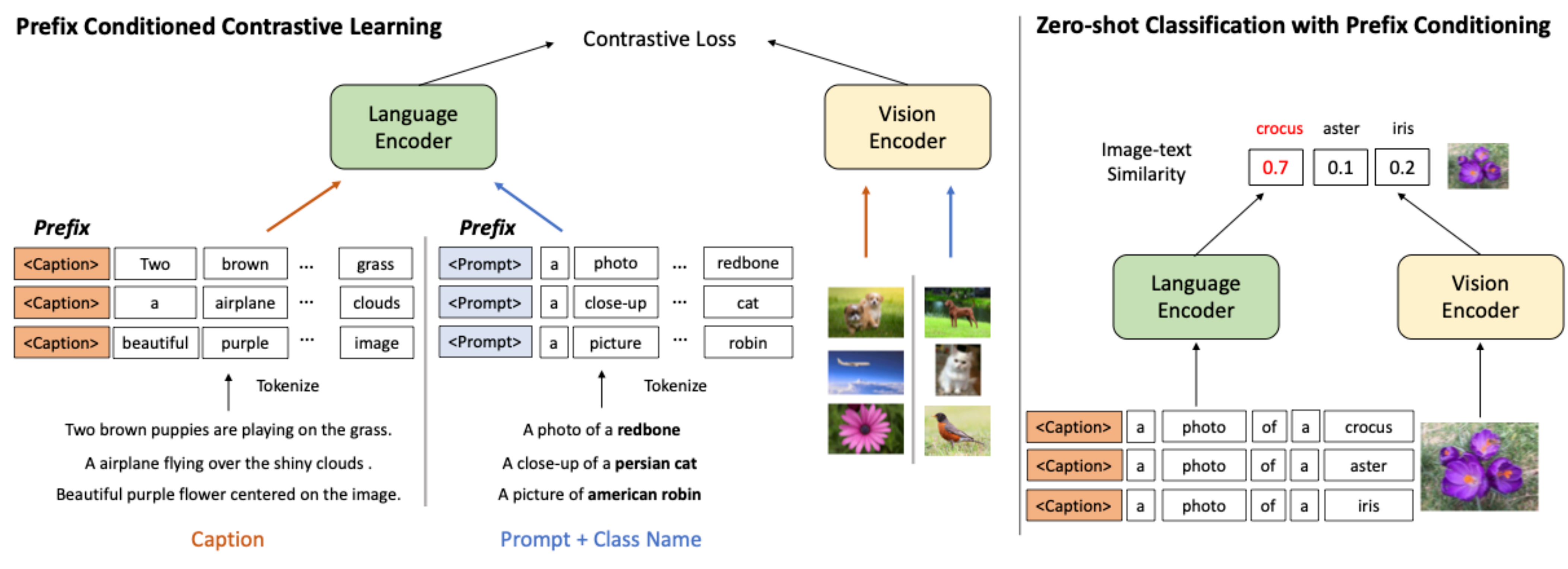}
     \vspace{-7mm}
    \caption{\textbf{Left}: Prefix conditioning at training time. Dataset-specific token is added to the input tokens with a contrastive learning objective applied. \textbf{Right}: Prefix conditioning at test time. Given a class name, we construct a class prompt with pre-defined templates and add a token used to condition real caption during training considering that image-caption dataset covers much wider range of image domains and vocabulary words than image classification dataset.}
    \label{fig:main_figure}
    \vspace{-3mm}
\end{figure*}
Prompt tuning is a popular technique to adapt a large language model to a specific task with few training data and low computational cost~\citep{li2021prefix,gao2020making,lester2021power,liu2021pre,qin2021learning}. To avoid tuning all parameters of the model and using hand-crafted prompts, prefix embeddings are added to the training input and are the only parameters optimized during fine-tuning. The prefix embedding can be viewed as the knowledge of the downstream task. In this paper, since the target task is the zero-shot classification, the bias of the language embedding needs to be from the dataset covering a wide range of domains rather than a specific domain. Therefore, we choose to use the prefix embedding learned for image-caption dataset during test time. 
This technique is also effective in adapting a pre-trained vision-language model~\citep{zhou2021learning, zhou2022conditional} to few-shot classification by tuning the prompts of the language encoder to adapt to a downstream task. Additionally, prompt-tuning is effective in adapting a pre-trained vision model to a target task~\citep{jia2022visual}. While these works aim to tailor a large pre-trained model to a specific downstream task with a small amount of data or low computational cost, our goal here is to condition a model with the prefix during the pre-training stage by distinguishing between the image label and image caption data. This allows a model to effectively share the knowledge obtained from two different types of data sources.

\textbf{Dataset bias in image recognition.}
A large-scale image recognition dataset such as ImageNet~\citep{deng2009imagenet} is known to be biased towards a specific image domain. Therefore, a model trained on such a dataset shows vulnerability to the distribution shift, e.g., shift in object pose~\citep{barbu2019objectnet} and style of the images \citep{wang2019learning}. 
Nevertheless, \cite{wortsman2022robust, kumar2022fine} show that adapting only a linear layer on the pre-trained models can improve performance on the downstream tasks with distribution shifts. 
This indicates the importance of having a good  classifier on top of image encoders, such as linear classifiers generated by language encoders with preconditioning in our work. 
\cite{dubey2021adaptive} propose a method for domain generalization. They condition image recognition models with the domain embedding, which discriminates the input image domains, and demonstrate the importance of the domain-specific image classifier. 
Our prefix conditioning can be seen as an attempt to de-bias the linear classifier to obtain a domain-specific classifier and adapt it from the classification to the captioning domain. 
Also, \cite{lee2021learning, bahng2020learning} approach the dataset bias in image classification by de-biasing image representations. By contrast, we tackle the problem in the framework of vision-language learning, disentangle the dataset bias in the language embedding and utilize the classifier obtained by the caption domain. We note that while captioning datasets can also have data biases, they tend to be more open-domain than existing classification datasets.



\vspace{-1mm}
\section{Method}
\label{sec:method}
\vspace{-1mm}
In this section, we introduce the Prefix Conditioning technique for pretraining a deep learning model on both image-caption and image-label (classification) data. In Sec.~\ref{sec:preliminary}, we discuss our problem setting and the background of contrastive learning with image-caption data. In Sec.~\ref{sec:prefix-condition}, we explain the details of our training approach, and in Sec.~\ref{sec:inference} our inference procedure. 

\vspace{-1mm}
\subsection{Preliminaries}
\label{sec:preliminary}
\vspace{-1mm}
\textbf{Setup.}
Suppose we have access to two datasets: (i) an image label dataset $\Scal_{L}  = \{ (\xv_n, \tv_n^{P}, y_n) \}_{n=1}^{N_{L}}$, where $\xv \in \Xcal$  is the image and $\tv^{P} \in \Pcal$ is a prompt-style language description based on its class label $y \in \Ycal$, and (ii) a dataset of image-caption pairs  $\Scal_{C}  =\{ (\xv_n, \tv_n^{C}) \}_{n=1}^{N_{C}}$, where $\tv^{C} \in \Tcal$ is a caption. We assume that $\tv$ is the tokenized language description.
For each image $\xv$, an image encoder model $f_{\thetav}$ parameterized by $\thetav$ extracts a visual representation $\Tilde{\vv}  \in \R^{d\times 1}$: $ \Tilde{\vv} = f_{\thetav}(\xv)$. For each caption or prompt  $\tv \in \Tcal$, a text encoder $f_{\phiv}$ parameterized by $\phiv$ extracts a language representation $ \Tilde{\uv}  \in \R^{d \times 1}: \Tilde{\uv}  = f_{\phiv}(\tv)$. 

\textbf{Contrastive Loss.}
CLIP~\citep{radford2021learning} is designed to find representations that match an image to its paired caption while separating unpaired ones.
For $i$-th image $\xv_i$ and $j$-th language description $\tv_j$ in a batch $\Bcal$, their features are normalized using $ \vv_i\,{ =}\, \frac{  \Tilde{\vv}_i   }{ \| \Tilde{\vv}_i \| }  $ and $ \uv_j \,{=}\, \frac{    \Tilde{\uv}_j }{   \| \Tilde{\uv}_j  \|} $. Finally, CLIP optimizes the symmetric multi-class N-pair loss \citep{sohn2016improved}:
\begin{align}\label{eq:obj_bicon}
	\min_{ \{ \thetav, \phiv \} } ~~ \Lcal_{con} 	= & \Lcal_{t2i} + \Lcal_{i2t}, 
\end{align}
which includes two contrastive terms (a temperature hyper-parameter $\tau$ controls the strength of penalties on hard negative samples):
\begin{align}
\Lcal_{t2i}	= - \frac{1}{ |\Bcal| }\sum_{ i \in \Bcal }  
\log \frac{ \exp(\tau \uv_{i}^T \vv_i)  }{\sum_{ j \in \Bcal}  \exp(\tau \uv_{i}^T \vv_{j})  },\\ \quad \Lcal_{i2t}	= - \frac{1}{ |\Bcal| }\sum_{ i \in \Bcal } 
\log \frac{ \exp(\tau \vv_{i}^T \uv_i )  }{\sum_{j \in \Bcal}  \exp(\tau \vv_{i}^T \uv_{j} )  }.
\end{align}
UniCL~\citep{yang2022unified} composes each mini-batch with  samples from both $\Scal_{L}$ and $\Scal_{C}$. Then, for pairs from $\Scal_{L}$, they regard all samples from the same class as positive pairs while a sample from $\Scal_{C}$ has a unique pair. Except for the number of positive pairs, no special treatment is given to differentiate between the image-caption and image-label data.

\subsection{Prefix Conditioned Contrastive Learning}
\vspace{-1mm}
\label{sec:prefix-condition}
Fig.~\ref{fig:main_figure} describes the overview of our approach. 
We aim to enable the language encoder to learn embedding strategies conditioned on the type of input dataset. The conditioning can then be used to manipulate the bias at inference time.

Prefix-tuning~\citep{li2021prefix,gao2020making,lester2021power,liu2021pre,qin2021learning} shares the intuition that the prefix tokens are responsible for switching the context of a language model from the pre-trained task to the downstream task. These approaches leverage the prefix to tailor a model to a single task during fine-tuning and construct different prefixes for different natural language tasks~\citep{lester2021power}. In our problem setting, there is no task distinction between the image-caption and image-prompt matching since both are formulated as contrastive learning. However, we focus on the fact that the two datasets have different biases in the image distributions and vocabulary words. The label-prompt sentences are embedded closer to the image classification data, even though we may want to use them to match a new label to an image from the open-domain image distribution during zero-shot classification. 

To solve this problem, we propose to inform the model of the type of dataset at the input level to switch the feature extraction. 
Specifically, to make the model aware of the dataset type, prefix-conditioning prepends a prefix token to an input sentence to obtain $\bar{\tv}^{P} = [\text{PREFIX}^{P}; \tv^{P}]$, $\bar{\tv}^{C} = [\text{PREFIX}^{C}; \tv^{C}]$. The brackets indicate the concatenation of two lists of discrete tokens;  $\text{PREFIX}^{P}$ and $\text{PREFIX}^{C}$ denote a prompt-style and caption-style token respectively. In this way, we prepend the token to learn the dataset-specific bias, which enables us to disentangle the bias in language representations and utilize the embedding learned on the image-caption dataset at test time \textit{even without an input caption}. 

In prompt-tuning, the number of prefix tokens can affect the performance of the model~\citep{lester2021power,zhou2021learning,li2021prefix}. However, we do not see the performance difference by the number of prefix tokens. This is probably because adding one token is enough to distinguish the domain of input sentences. To avoid significantly increasing the training cost, we set the number of prefixes to one in all experiments. 
Then, the language representations for each data source are extracted as $\tilde{\uv}^{P}, \tilde{\uv}^{C}  \,{=}\, f_{\phiv}(\bar{\tv}^{P}), f_{\phiv}(\bar{\tv}^{C})$. This input design is independent from the training objectives, and therefore we can easily apply the technique to optimize Eq.~\ref{eq:obj_bicon} or UniCL's loss.

\textbf{Data Sampling.} 
\cite{cui2021zerovl} argue that the data sampling matters when learning from multiple data sources in a contrastive learning framework, as the model may learn to distinguish the samples by exploiting the dataset bias. As such, we need to take data sampling into consideration in our problem setting as we learn from two different data sources. 
One option is a debiased sampling~\citep{cui2021zerovl}, which constructs each mini batch to contain samples from a single data source. Alternatively, as done in UniCL~\citep{yang2022unified}, we can compose each mini-batch with samples from both data sources (image-caption and image-label) with equal probability. In experiments, we choose the debiased sampling, but find that the choice of sampling does not significantly affect the performance.

\subsection{Inference with Prefix Conditioning}\label{sec:inference}
\vspace{-1mm}
During inference (the right side of Fig.~\ref{fig:main_figure}), an input image is classified as one of $K$ classes by embedding the corresponding label-prompts and choosing the one most similar to the image embedding. 
 Following~\citep{radford2021learning}, we obtain class prompts by filling the default prompt templates with class names, and add a prefix. Considering the wider coverage of domains in the image-caption dataset, the caption-style prefix conditioning may work better to classify novel downstream data.
In our experiments, we empirically find that the caption-style prefix indeed outperforms the prompt-style prefix with a large margin in zero-shot recognition while prompt prefix performs better on the image classification dataset used to train the model.
We provide a detailed analysis of different conditioning in Section~\ref{sec:analysis}.

\begin{table*}[t]
    \centering
    \resizebox{0.75\linewidth}{!}{%
    \begin{tabular}{llc|c|c|cc}
    \toprule
     \multicolumn{3}{c|}{\multirow{2}{*}{Training Data}} &  \multirow{3}{*}{Objective} & \multirow{3}{*}{\makecell{Prefix \\Conditioning}} &  \multicolumn{2}{c}{Metric}  \\
        \cmidrule{6-7}
         \multirow{2}{*}{Classification} & \multirow{2}{*}{Caption} &\multirow{2}{*}{Size}   &&& \multirow{2}{*}{IN-1K} & \multirow{2}{*}{\makecell{Zero-shot \\ 11 datasets}}  \\
        & &  &&&&\\
        \midrule
        -- &CC-3M &3M& CLIP  &&18.1 & 28.7\\
        -- &CC-12M &12M& CLIP &&33.4 & 41.2\\
        ImageNet-1K & -- &1M& CLIP &&72.1&20.2\\
        ImageNet-21K & -- &12M& CLIP &  & 47.1 & 39.6\\
        
         \midrule\midrule
         ImageNet-1K & CC-12M &13M&CLIP &&  68.7&43.3\\
         ImageNet-1K & CC-12M &13M&CLIP &\checkmark& \bf{71.5}&\bf{45.5}\\
          \midrule
          ImageNet-1K & CC-12M &13M&UniCL &&  68.8&43.1\\
          ImageNet-1K & CC-12M &13M&UniCL &\checkmark& \bf{71.7} &\bf{44.5}\\
        \midrule\midrule
        ImageNet-21K & CC-12M  &25M&  CLIP & &56.8&49.5 \\
        ImageNet-21K & CC-12M  &25M&  CLIP &\checkmark&\bf{67.3}&\bf{57.8}\\ 
        \midrule
        ImageNet-21K & CC-12M  &25M&  UniCL & & 58.2&51.7\\
        ImageNet-21K & CC-12M  &25M&  UniCL  &\checkmark & \bf{66.5}&\bf{58.4}\\
        \midrule \midrule
         ImageNet-21K w/o IN-1K & CC-12M  &24M&  CLIP &&29.1&46.9\\ 
          ImageNet-21K w/o IN-1K & CC-12M  &24M&  CLIP &\checkmark&\bf{47.8}&\bf{56.4}\\ 
        \bottomrule
    \end{tabular}
    }
    \vspace{-3mm}
    \caption{Performance comparison among different training datasets and training objectives. Note that we use caption prefix to obtain these results. The proposed prefix conditioning shows improved zero-shot recognition accuracy across models trained with different combinations of image-classification and image-caption datasets and training objectives.}
        \label{tab:imagenet21K_plus_imagetext}
\end{table*}
\vspace{-3mm}
\section{Experiments}
\vspace{-1mm}
\label{sec:exp}
The goal of experiments is twofold: comparing our approach with baselines in zero-shot recognition, and  analyzing the behavior of prefix conditioning. We  describe the experimental setup in Sec.~\ref{sec:setup}, show the main results in Sec.~\ref{sec:main_results}, and analyze the properties of prefix-conditioning in Sec.~\ref{sec:analysis}. 
\vspace{-3mm}
\subsection{Setup}
\vspace{-1mm}
\label{sec:setup}
\textbf{Training Datasets.}
We conduct experiments on the setting where we have a large source of image-caption and image-label datasets. Following UniCL~\citep{yang2022unified}, we utilize CC3M~\citep{sharma2018conceptual} and CC12M~\citep{changpinyo2021cc12m} as image-caption data.
For the image classification dataset, we utilize ImageNet21K and ImageNet1K~\citep{deng2009imagenet}. While ImageNet1k contains 1,000 classes, ImageNet21K has more than 20,000 categories that include fine-grained and general objects. To observe the behavior in diverse image classification data, we also run experiments on ImageNet21K while excluding the classes of ImageNet1K. Details  are explained in each section.

\textbf{Training.} 
We use the same prompt strategy and 80 prompt templates as used in CLIP~\citep{radford2021learning}. During training, we
randomly sample one prompt template and fill it with the class
names, followed by a tokenization step
before feeding into the text encoder. We average language embeddings extracted from all 80 templates in validation. We use the same language encoder as CLIP~\citep{radford2021learning} and Swin-Tiny transformer ~\citep{liu2021Swin} as the vision encoder following UniCL~\citep{yang2022unified}. 
All models are optimized with AdamW~\citep{loshchilov2018decoupled} where the learning rate is set to 0.001, and weight decay to 0.1. 
All models are trained with a batch size of 1024. Considering the amount of training data, we train the models for 15 and 50 epochs in the experiments on ImageNet21K and ImageNet1K respectively.\footnote{When training a model on two different datasets, e.g., IN21K and CC12M, we count the epochs based on how many samples are used from the image classification dataset. For instance, in UniCL, each mini-batch consists of approximately the same number of samples from IN21K and CC12M. Then, to train a model for 15 epochs, we train for $N / 1024 \times 2 \times 15$ iterations, where $N$ indicates the number of  samples in IN21K.}
For all training, we used a cosine learning rate schedule with a warm-up of 10,000 iterations.

\textbf{Baselines.} 
We train CLIP~\citep{radford2021learning} and UniCL~\citep{yang2022unified} as our baselines. For comparison, we present results on CLIP trained only on image-caption or image classification data, as well as CLIP and UniCL trained on both image-caption and IN21K data. Unless otherwise stated, CLIP and UniCL are trained with equal sampling (ES) strategy as in \cite{yang2022unified}, while our prefix conditioning model is trained with debiased sampling (DS)~\citep{cui2021zerovl}. We provide an analysis of the sampling in Sec.~\ref{sec:main_results} and find that DS itself does not have a noticeable advantage over ES.

\textbf{Evaluation.} 
We evaluate the learned representations on supervised and zero-shot image classification on ImageNet1K\footnote{While we follow the same zero-shot evaluation protocol when evaluating on ImageNet1K, we note that it is zero-shot only where we explicitly exclude ImageNet1K from the training, last two rows of Table 1} and on 11 datasets chosen from the ones used in CLIP~\citep{radford2021learning} including object classification (e.g., CIFAR10, CIFAR100), fine-grained classification (e.g., Oxford-IIIT Pets, Oxford Flowers 102, and  Food-101), and aerial images (e.g., EuroSAT and Resisc45). Although our main focus is at the zero-shot generalization, we also provide an analysis of a linear-probe evaluation of the image encoder.

\begin{table*}[t]
    \centering
    \resizebox{0.99\linewidth}{!}{%
    \begin{tabular}{c|c|cccccccccccc|c}
    \toprule
    \multirow{2}{*}{\makecell{Train \\ Prefix}} & \multirow{2}{*}{Sampling}& \multirow{2}{*}{IN-1K} & \multirow{2}{*}{Cal}	& \multirow{2}{*}{CF100}&\multirow{2}{*}{CF10}&\multirow{2}{*}{ESTAT}&\multirow{2}{*}{Food}&\multirow{2}{*}{Flower}&\multirow{2}{*}{Pets}&\multirow{2}{*}{Patch}&\multirow{2}{*}{R45}&\multirow{2}{*}{VOC}&\multirow{2}{*}{DTD}&\multirow{2}{*}{AVG}\\
    &&& &&&&&&&&&&\\
      \cmidrule{1-15}
         &ES&56.8&70.2&55.0&79.4&21.1&46.0&60.3&57.2&51.2&24.8&57.7&21.4&49.5\\
         \checkmark&ES&\bf{65.4}&\bf{81.2}&\bf{62.6}&\bf{88.9}&\bf{30.4}&\bf{51.7}&\bf{61.8}&\bf{71.9}&50.0&\bf{28.2}&\bf{78.1}&	\bf{27.7}&\bf{57.5}\\
         \midrule
         &DS&58.7&65.9&55.0&85.7&22.8&40.8&55.7&60.2&50.0&20.6&	45.2&23.8&47.8\\
        \checkmark&DS&\bf{67.3}&\bf{79.7}&\bf{63.8}&\bf{87.9}&\bf{31.5}&	\bf{53.4}&\bf{58.8}&\bf{69.6}&\bf{50.6}&\bf{31.5}&\bf{80.5}&\bf{28.4}&\bf{57.8}\\
        \bottomrule
    \end{tabular}
    }
     \vspace{-3mm}
     \caption{Ablation study for sampling in IN21K + CC12M. Equal sampling (ES) composes a mini-batch with roughly equal number of samples from two datasets. Debiased sampling (DS) samples a mini-batch of either IN21K or CC12M with equal probability.}
    \label{tab:sampling_analysis}
     \vspace{-2mm}
\end{table*}

\begin{table*}[!t]
    \centering
   \resizebox{0.8\linewidth}{!}{%
    \setlength{\tabcolsep}{2.5pt}
    \begin{tabular}{c|c|ccccccccc}
    \toprule
    \multirow{2}{*}{Train Data}& \multirow{2}{*}{\makecell{Prefix \\ Conditioning}}& \multirow{2}{*}{IN-1K} & \multirow{2}{*}{Cifar10}& \multirow{2}{*}{Cifar100}&\multirow{2}{*}{Caltech}&
    \multirow{2}{*}{Food}&\multirow{2}{*}{Pet}&\multirow{2}{*}{Patch}&\multirow{2}{*}{VOC}&\multirow{2}{*}{DTD}\\
    &\\
     \midrule
    ImageNet-21K&&71.5&94.3&79.1&83.5&79.1&86.3&82.3&88.9&61.3 \\
    ImageNet-21K + CC12M&&69.2&93.0&76.4&82.4&78.4&82.2&81.4&88.7&61.4\\
    ImageNet-21K + CC12M&\checkmark&69.4&93.5&77.3&83.2&78.8&83.6&82.0&88.8&62.5\\
        \bottomrule
    \end{tabular}}
     \vspace{-3mm}
     \caption{Linear evaluation accuracy on models trained with and without prefix conditioning. Prefix conditioning slightly improves the performance upon a model without it (second row vs. last row).}
    \label{tab:linear_evaluation}
\end{table*}

\vspace{-2mm}
\subsection{Main Results}
\vspace{-2mm}
\label{sec:main_results}
We describe our main results in Table~\ref{tab:imagenet21K_plus_imagetext}, followed by the analysis of prefix conditioning in Sec.~\ref{sec:analysis}.

There are three observations. First, the improvements upon a model trained only with image-caption or image-label data are obvious in almost all cases. As the previous work indicates~\citep{yang2022unified}, the effectiveness of combining two types of supervision is clear from these results. 

Second, in all cases, our prefix conditioning significantly improves performance on both ImageNet-1K (supervised recognition) and 11 zero-shot recognition tasks. When training on ImageNet-21K, the conditioning improves the baseline by more than 8\% in ImageNet-1K and more than 6\% in zero-shot recognition on average. In training with ImageNet-1K, the margin from the baseline is smaller than training with ImageNet-21K, probably because the size of ImageNet-1K (1M) is much smaller than that of ImageNet-21K (12M). Also, prefix conditioning is effective in both UniCL and CLIP objectives. Due to its simplicity, our approach can be easily integrated with various objectives.  

Finally, our method is less affected by ablating a part of categories. The classes of ImageNet-1K are excluded from ImageNet-21K in the last two rows of Table~\ref{tab:imagenet21K_plus_imagetext}. Therefore, both approaches significantly drop performance on ImageNet-1K, whose task now becomes true zero-shot recognition, compared to other settings. Even in this setting, prefix conditioning maintains high accuracy and outperforms a CLIP baseline model by a large margin. 

\textbf{Sampling Method.}
We analyze the data sampling scheme to construct a mini-batch in Table~\ref{tab:sampling_analysis}.
We apply debiased sampling (DS) in our method, namely, sampling one data source with equal probability and getting a mini-batch of it. The other option is mixing two data sources with equal probability (ES). The table indicates that prefix conditioning works well with ES sampling and the sampling strategy itself is not advantageous. Ablating prefix conditioning during training clearly drops the performance in both sampling strategies, and the performance is worse than ES on average in zero-shot results (49.5 vs. 47.8). ES sampling should allow the model to differentiate sentences by using the prepended prefix. Interestingly, this result implies that differentiating sentences by prefix information does not much degrade the performance. The distinguished sentences enable the model to associate images from different datasets. Since images of two datasets are different with respect to the categories and the locations of objects in images, distinguishing the two kinds of images may not harm generalizability of the representations.

\textbf{Linear-probe Evaluation.}
We evaluate the linear-probe performance in Table~\ref{tab:linear_evaluation} to see the quality of learned image representations. Although the accuracy is better than the model trained without prefix conditioning (second line), the improvements are not substantial. This result indicates that the zero-shot performance gain obtained by our method is not due to the image representations. We investigate the learned language and image features in the next subsection.
\begin{table*}[!t]
    \centering
    \resizebox{0.99\linewidth}{!}{%
    \begin{tabular}{c|c|cccccccccccc|c}
    \toprule
    \multirow{2}{*}{\makecell{Data}}& \multirow{2}{*}{\makecell{Test-time \\ Prefix}}& \multirow{2}{*}{IN-1K} & \multirow{2}{*}{Cal}	& \multirow{2}{*}{C100}&\multirow{2}{*}{C10}&\multirow{2}{*}{ESTAT}&\multirow{2}{*}{Food}&\multirow{2}{*}{Flower}&\multirow{2}{*}{Pets}&\multirow{2}{*}{Patch}&\multirow{2}{*}{R45}&\multirow{2}{*}{VOC}&\multirow{2}{*}{DTD}&\multirow{2}{*}{AVG}\\
    &&&&&&&&&&&&&& \\
      \cmidrule{1-15}
    \multirow{3}{*}{\makecell{IN-1K \\+ CC12M}}&N/A&68.7&68.7&38.4&69.5&24.4&31.9&13.3&66.6&	50.2&25.4&65.6&22.3&43.3\\
    &Prompt&\bf{75.4}&71.7&35.5&63.9&24.2&20.0&8.1&72.2&50.4&24.2&61.1&15.3&40.6\\ 
    &Caption&71.5&\bf{75.1}&\bf{39.4}&\bf{70.5}&\bf{26.7}&\bf{33.9}&\bf{13.9}&\bf{72.3}&\bf{50.5}&\bf{25.8}&\bf{67.8}&\bf{25.4}&\bf{45.5}\\
     \midrule 
\multirow{3}{*}{\makecell{IN-21K \\+ CC12M}}&N/A&56.8&70.2&55.0&79.4&21.1&46.0&60.3&57.2&51.2&	24.8&57.7&21.4&49.5\\
&Prompt&\bf{71.4}&76.5&59.0&86.0&20.1&45.7&\bf{62.3}&69.1&\bf{52.4}&26.3&76.8&21.4&54.1\\
 & Caption&67.3&\bf{79.7}&\bf{63.8}&\bf{87.9}&\bf{31.5}&	\bf{53.4}&58.8&\bf{69.6}&50.6&\bf{31.5}&\bf{80.5}&\bf{28.4}&\bf{57.8}\\
        \midrule 
 \multirow{3}{*}{\makecell{IN-21K w/o 1K\\+ CC12M}} &N/A &29.1&67.4&45.9&80.0&28.6&40.8&56.9&	39.2&50.2&21.9&64.9&19.8&46.9\\
 &Prompt & 40.8&74.9&61.0&84.6&31.2&48.1&58.7&45.2&\bf{51.2}&	23.5&67.5&21.4&51.6\\
&Caption&\bf{47.8}&\bf{81.9}&\bf{63.3}&\bf{87.3}&\bf{32.4}&\bf{52.9}&\bf{62.8}&\bf{57.0}&50.6&\bf{25.6}&\bf{80.1}&\bf{26.2}&\bf{56.4}\\
        \bottomrule
    \end{tabular}
    }
     \vspace{-3mm}
     \caption{Ablation study for test-time prefix conditioning. Note that the difference between two results come from the prefix used in test time and we use the same model for this evaluation. A model trained without conditioning is shown at the top of each block.}
    \label{tab:prefix_analysis}
     \vspace{-3mm}
\end{table*}

\begin{figure*}
    \centering
    \includegraphics[width=\textwidth]{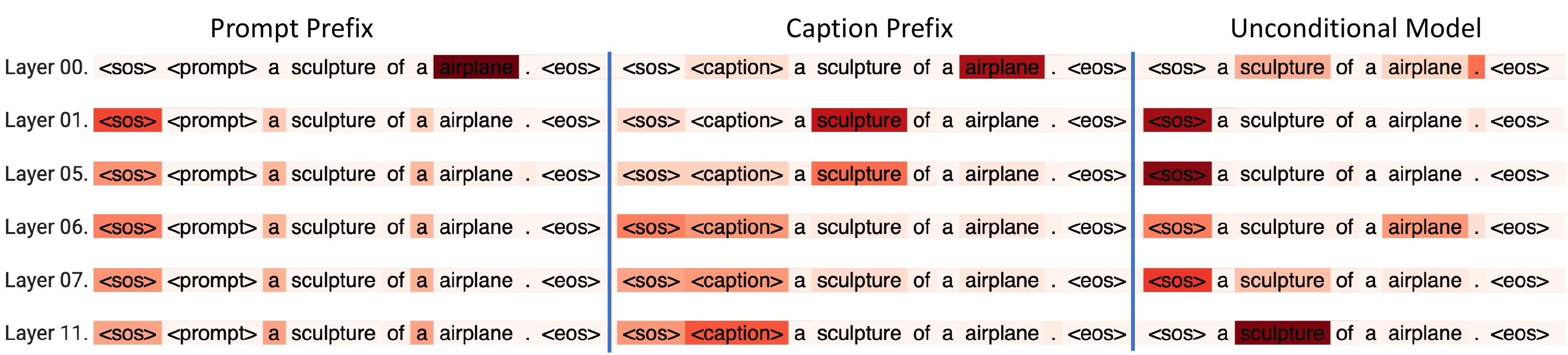}
    \vspace{-7mm}
    \caption{An example of attention weights for an end token. Best viewed in color. The sentence shown here is one of class prompts in the VOC 2007 dataset.  Different rows show the weights of different transformer layers. With a prompt prefix (leftmost), the model focuses on a class name (\textit{airplane}) while caption prefix (middle) allows a model to pay attention to another noun, \textit{sculpture}. By prefix conditioning, the attention of the model changes as intended.}
    \label{fig:attention_visualize}
\end{figure*}

\vspace{-2mm}
\subsection{Analysis of Prefix-Conditioning}
\vspace{-2mm}
We present a detailed analysis of prefix conditioning. We first study how different prefixes impact the zero-shot recognition performance and analyze their behaviors by looking into the attention weights of the language transformer encoder. We also demonstrate improved robustness with respect to the image-level domain shift. Unless otherwise stated, we employ a model trained with CLIP objective on ImageNet-21K and CC12M in this analysis.
%
%
Finally, this section concludes that prefix conditioning enables the language encoder to switch its role during training, which eases learning from different types of datasets, e.g., image classification and image caption dataset.

\label{sec:analysis}
\textbf{Test Time Prefix.}
We analyze the role of the prefix token in Table~\ref{tab:prefix_analysis}, where the table describes the comparison in the choice of test time prefix conditioning. 
As explained in Sec.~\ref{sec:method}, the choice of prefix during test time should change the behavior of the model since the prefix should tailor the language encoder for classification-style or caption-style feature extraction. 
%
Except for the IN-1K results of a model trained with the entire IN21K or IN-1K, conditioning with the caption prefix shows much better results. 
The superiority of the caption prefix is noticeable in several datasets. 
This means caption prefix works better if the target comes from outside the image classification data, indicating that the class-prompt prefix conditioning makes the model tailored for the image classification dataset. Class-prompts prefix works better to categorize IN-1K data because the prefix is trained to specialize in classifying it. 
Note that caption-style prefix performs better than prompt-style prefix in IN-1K for a model trained with IN21K excluding IN1k classes. This indicates that the caption-style prefix works better when the vocabulary of the class name comes from outside the image classification data since the caption data covers much more diverse words.

\textbf{Prefix controls attention.}
Fig.~\ref{fig:attention_visualize} visualizes the attention weights for an end token in different prefix conditions and models. The input sentence, \textit{a sculpture of an airplane}, is one of the class-prompts. 
When a prompt prefix (leftmost) is employed, the language model pays attention to the class name at the first layer, it does not focus on the noun in other layers. The only noun the encoder focuses on is \textit{airplane}. By contrast, the model attends to both \textit{sculpture} and \textit{airplane} in the case of the caption prefix and unconditional model. Note that this behavior does not mean that the prompt-prefix performs better in zero-shot recognition as shown in experiments due to the effect of the bias in image classification dataset.

\begin{table*}[t]
    \centering
     \resizebox{0.65\linewidth}{!}{%
    \begin{tabular}{c|c|c|ccccc}
    \toprule
    \multirow{2}{*}{Train Data}&\multirow{2}{*}{\makecell{Prefix \\ Conditioning}}&\multirow{2}{*}{\makecell{Test-Time \\ Prefix}}& \multirow{2}{*}{IN} &  \multirow{2}{*}{IN-V2} & \multirow{2}{*}{IN-R}&\multirow{2}{*}{IN-S}\\
    &&&&&&\\
     \midrule
    ImageNet-1K&&N/A&72.1&59.3&19.9&17.8\\
    ImageNet-1K + CC12M&&N/A&68.7&57.4&27.7&27.8\\
    ImageNet-1K + CC12M&\checkmark&Caption&71.5&60.2&\bf{31.8}&\bf{30.7}\\
    ImageNet-1K + CC12M&\checkmark&Prompt&\bf{75.4}&\bf{63.3}&29.2&27.9\\
    \midrule
    
    ImageNet-21K&&N/A&47.1&41.1&20.1&16.1\\
    ImageNet-21K + CC12M&&N/A&56.8&48.6&29.4&30.6\\
    ImageNet-21K + CC12M&\checkmark&Caption&67.3&57.5&\bf{35.2}&\bf{34.6}\\
    ImageNet-21K + CC12M&\checkmark&Prompt&\bf{71.4}&\bf{61.1}&32.1&32.2\\
        \bottomrule
    \end{tabular}}
     \vspace{-3mm}
    \caption{Evaluation on the robustness to the image-level domain shift. Prefix conditioned training achieves better robustness, and caption-prefix outperforms prompt-prefix in the images distinct from those used in training (IN-R and IN-S). }
    \label{tab:robustness_evaluation}
\end{table*}

While we visualize only one example in the main text due to the space limit and defer more examples to the appendix, this highlights a general trend that the prompt prefix guides the language encoder to focus on a single word (e.g., class name), whereas the caption prefix makes the model attend to multiple words.
In other words, prefix conditioning allows the language encoder to ``switch gears'' to represent sentences from different datasets (i.e., image-classification vs image-caption). On the other hand, the baseline model without prefix conditioning attends to multiple words (e.g., Fig.~\ref{fig:attention_visualize} rightmost) even though the input sentence is a class prompt. This indicates that it is hard to switch the gears without explicitly informing of the type of dataset.
\begin{figure}
     \centering
     \begin{subfigure}[b]{0.23\textwidth}
         \centering
         \includegraphics[width=\textwidth]{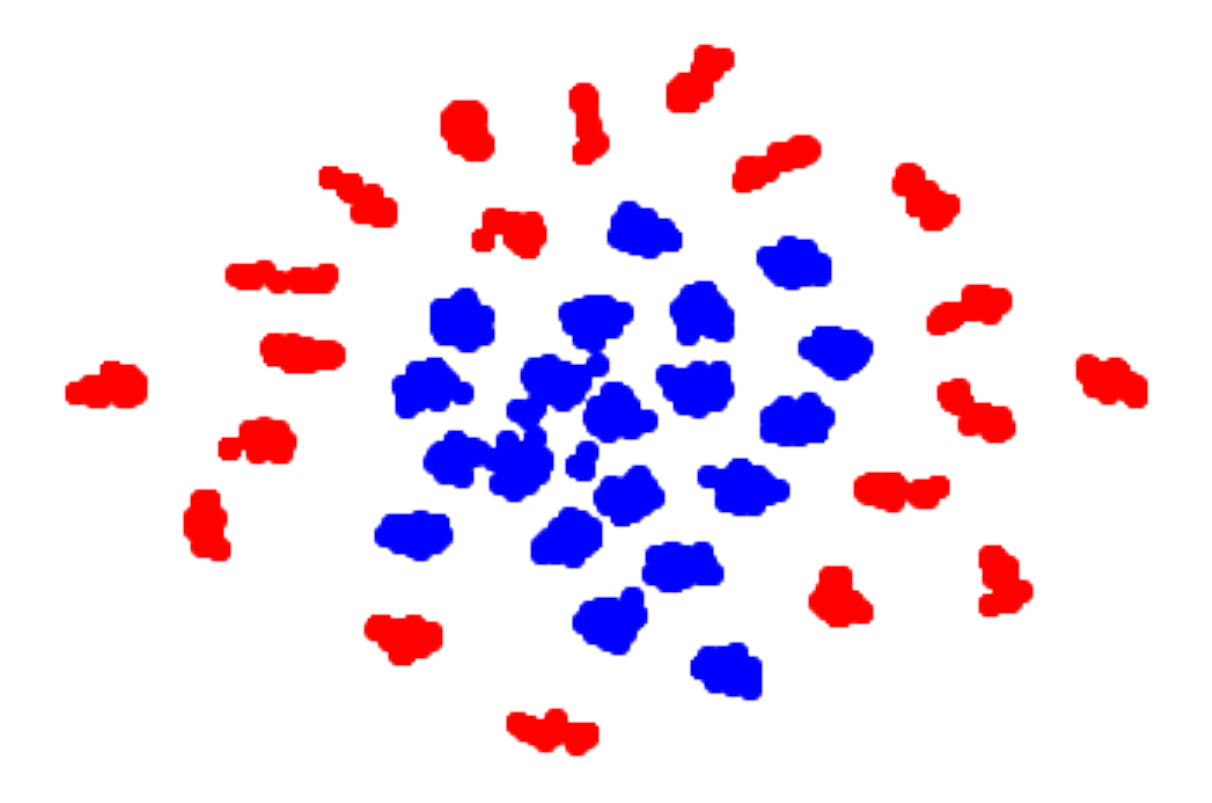}
         \caption{Different conditions}
         \label{fig:domain_plot}
     \end{subfigure}
     \begin{subfigure}[b]{0.23\textwidth}
         \centering
         \includegraphics[width=\textwidth]{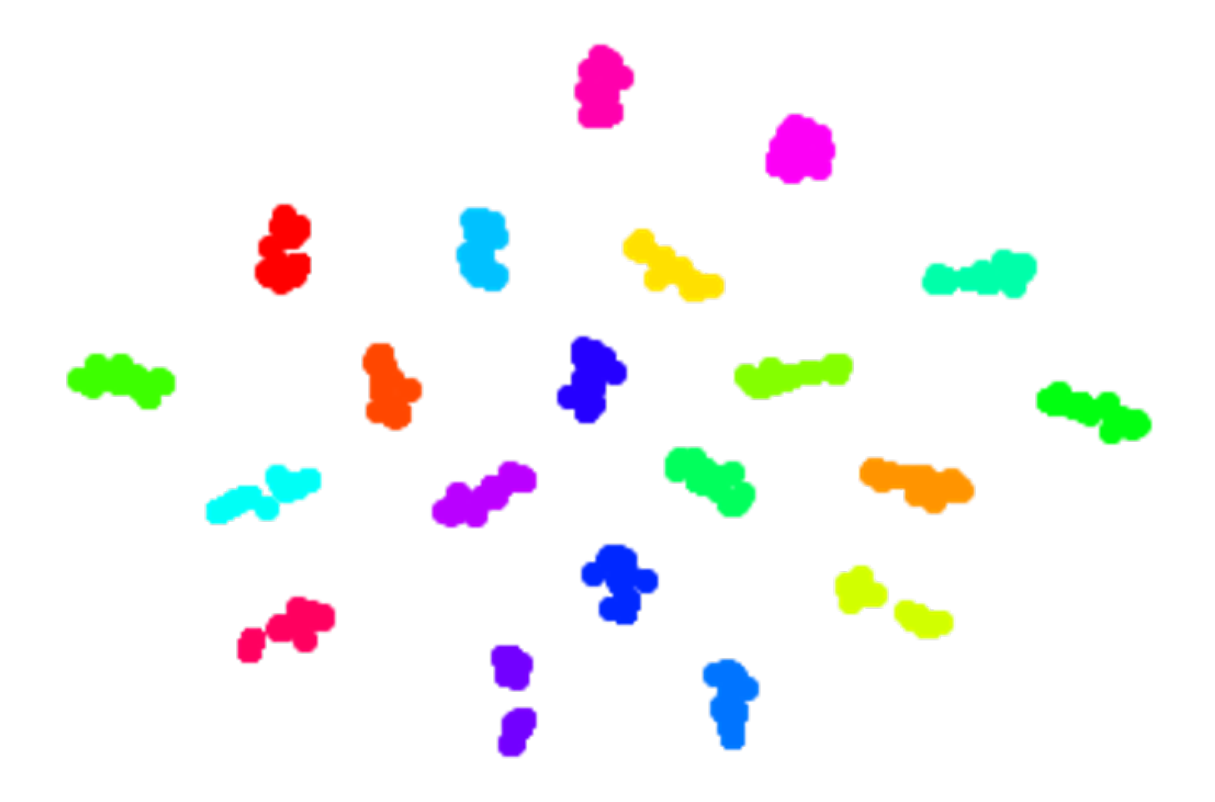}
         \caption{Prompt condition}
         \label{fig:class_plots}
     \end{subfigure}
     \begin{subfigure}[b]{0.23\textwidth}
         \centering
         \includegraphics[width=\textwidth]{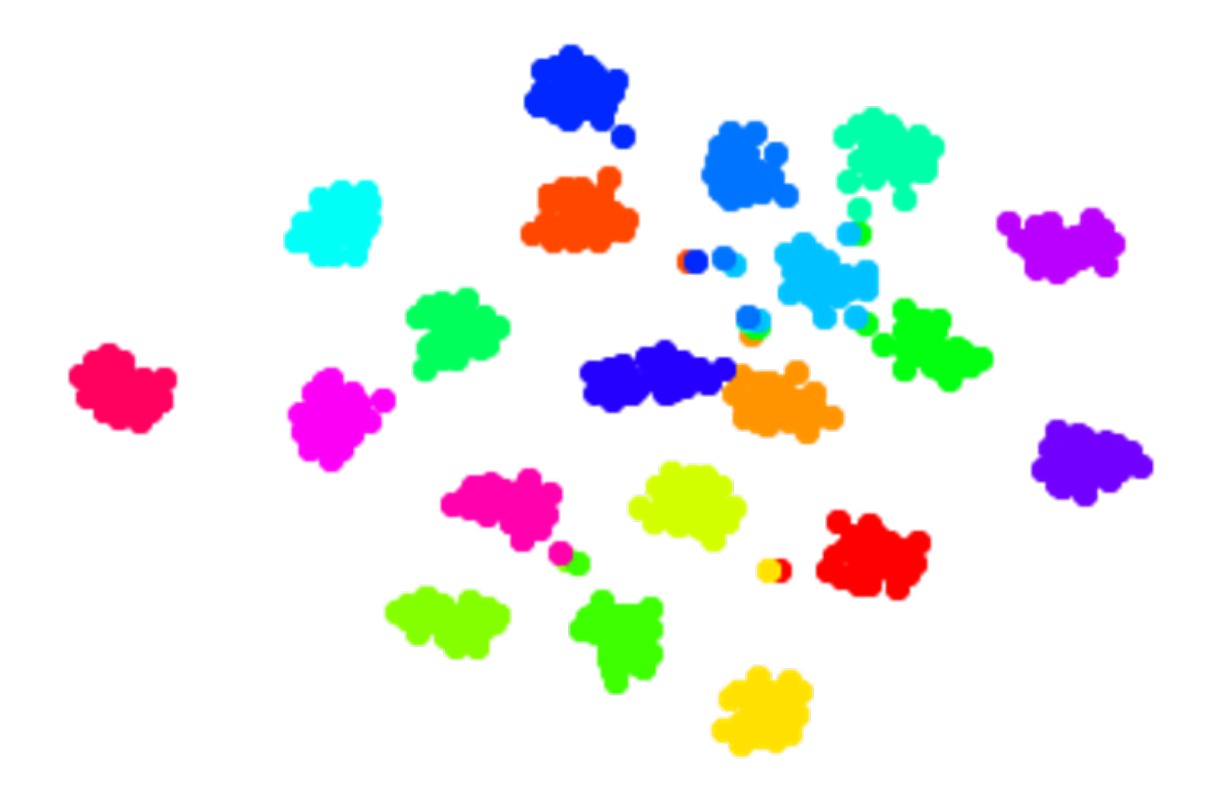}
         \caption{Caption condition}
         \label{fig:caption_plots}
     \end{subfigure}
     \begin{subfigure}[b]{0.23\textwidth}
         \centering
         \includegraphics[width=\textwidth]{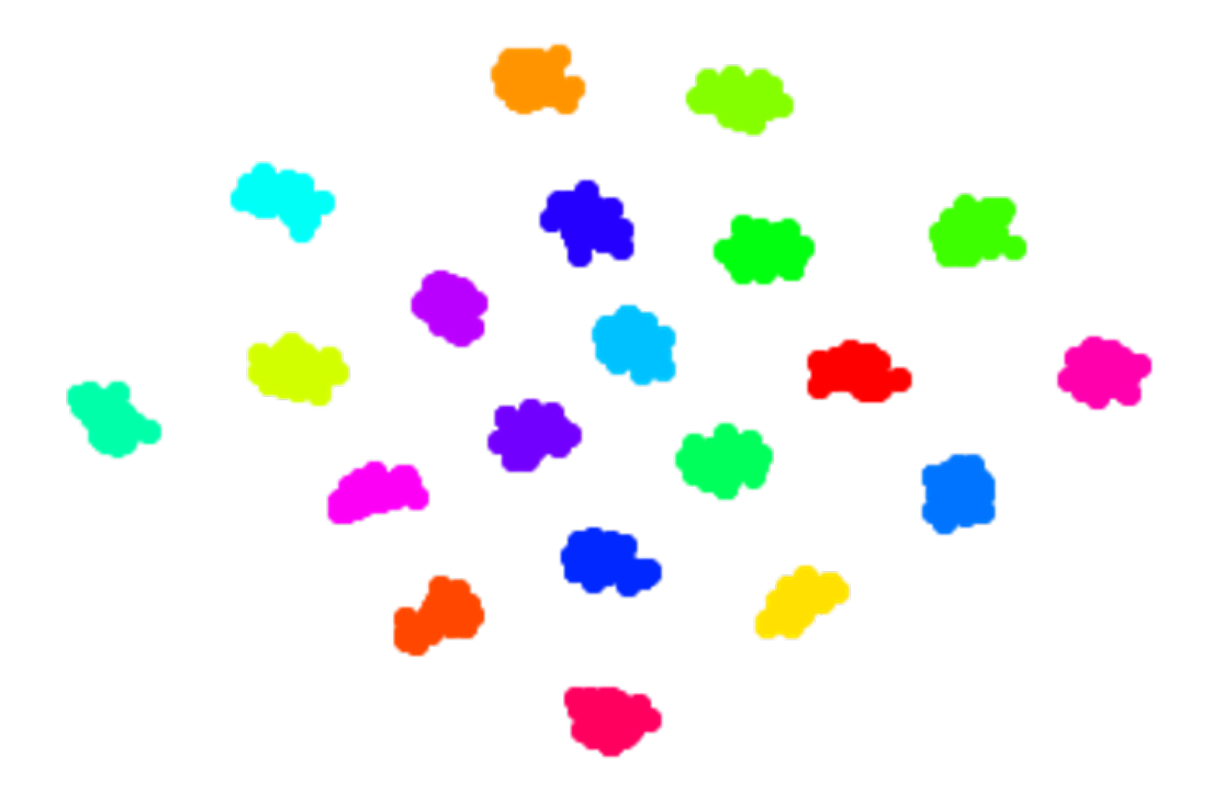}
         \caption{No condition}
         \label{fig:uncondition_plots}
     \end{subfigure}
    \vspace{-3mm}
      \caption{T-SNE~\citep{van2008visualizing} visualization of the class-prompt features of 20 classes of VOC 2007 with different prefix conditions. (a): Language embeddings with prompt (red) and caption (blue) prefixes, respectively. (b)(c)(d): Different colors indicate language embeddings of different classes. }
      \label{fig:tsne_plots}
\end{figure}

\textbf{Language Feature Visualization.}
Fig.~\ref{fig:tsne_plots} visualizes extracted language features conditioned with different prefixes. As seen in Fig.~\ref{fig:domain_plot}, language features extracted with caption-prefix (blue) and prompt-prefix (red) are clearly separated. In addition, prompt-prefix (Fig.~\ref{fig:class_plots}) has lower intra-class and higher inter-class variance, whereas caption-prefix (Fig.~\ref{fig:caption_plots}) shows higher intra-class variance across prompts.
Interestingly, results in Table~\ref{tab:prefix_analysis} suggest that the caption-prefix conditioned language features result in a better zero-shot recognition performance than those conditioned on the prompt-prefix. Although the prompt-prefix mode extracts discriminative language embeddings, the embeddings do not perform well on the zero-shot recognition because the embeddings contain significant bias from image-classification dataset. 
%


\textbf{Robustness in image domain shift.}
Test samples can be unseen with respect to image classification data in two ways (or combinations of two): 1) The image is similar to the training distribution, but the class name is different from the seen image classification labels. 2) Although the class label is the same, the image data comes from a different distribution. Datasets evaluated in the zero-shot recognition include both two cases since the vocabularies and image are from different domains. To understand them, we analyze the test-time prefix by using ImageNet-1K and evaluate the performance on image-level domain shift using variants of ImageNet, i.e., ImageNet-V2~\citep{recht2019imagenet}, ImageNet-R~\citep{hendrycks2021many}, and ImageNet-S~\citep{wang2019learning}.  Table~\ref{tab:robustness_evaluation} describes the results of ablating prefix-conditioned training and the test-time prefix. The prefix-conditioned training outperforms all baselines. This reveals that the prefix-conditioned training achieves class embeddings that are generalizable across image domains. The prompt-style prefix performs the best in IN, IN-V2, both of which have image styles similar to ImageNet. By contrast, the caption-style prefix performs the best in IN-R and IN-S, which has art-style and sketch-style images respectively. Thus, the caption-style prefix generates more generalizable class embeddings for the domain dissimilar from the ImageNet training data. This observation is consistent with the results in the paragraph \textit{Test time Prefix}.



\vspace{-3mm}
\section{Conclusion}
\vspace{-1mm}
In this paper, we explore a simple yet effective mechanism for unified pre-training on image-caption and image classification data. We propose to learn prefix tokens at training time to condition the language encoder to switch the input source. 
Specifying the prefix allows the model to switch the manner of feature extraction and can control which visual domain the embedding is projected to.
This approach boosts the performance of zero-shot recognition accuracy of the contrastive learning models. Our analysis suggests that the trained language encoder provides robustness to the image-level domain shift. Although we limit our scope to unifying image-caption and image-label supervision, incorporating other supervision such as object detection or semantic segmentation is an interesting next step. 


\label{sec:conc}
\noindent \textbf{Acknowledgment.}
We thank Zizhao Zhang for their helpful feedback on the manuscript. This work was supported in part by DARPA LwLL.

\appendix

\begin{table*}[h!]
  \centering
  \scalebox{0.99}{
\begin{tabular}{cc|cccccc} 
 \toprule
Abbreviation & Dataset & \#Concepts  & Train size & Test size &Source link \\ 
 \midrule
 Food&Food-101 & 102 & 75,750 & 25,250 & \href{https://www.tensorflow.org/datasets/catalog/food101}{Tensorflow}\\
CF10& CIFAR-10 & 10 &  50,000 & 10,000  &\href{https://www.tensorflow.org/datasets/catalog/cifar10}{Tensorflow}  \\
CF100& CIFAR-100 & 100 & 50,000 & 10,000 &\href{https://www.tensorflow.org/datasets/catalog/cifar100}{Tensorflow}  \\
VOC&VOC2007 classification & 20 &  5,011 & 4,952& \href{https://www.tensorflow.org/datasets/catalog/voc}{Tensorflow}    \\
DTD&Describable Textures & 47 & 3,760& 1,880&\href{https://www.tensorflow.org/datasets/catalog/dtd}{Tensorflow} \\
Pets&Oxford-IIIT  Pets & 37  & 3,680& 3,669 &\href{https://www.tensorflow.org/datasets/catalog/oxford_iiit_pet}{Tensorflow} \\
Cal&Caltech-101& 102 &  3,060& 6084 &\href{https://www.tensorflow.org/datasets/catalog/caltech101}{Tensorflow}  \\
Flower&Oxford Flowers 102& 102 & 1,020& 6,149 &\href{https://www.tensorflow.org/datasets/catalog/oxford_flowers102}{Tensorflow} \\
Patch&PatchCamelyon & 2  & 294,912 & 32,768 &\href{https://www.tensorflow.org/datasets/catalog/patch_camelyon}{Tensorflow}  \\
ESTAT&EuroSAT &	10	  & N/A  & 27,000 &\href{https://www.tensorflow.org/datasets/catalog/eurosat}{Tensorflow} \\
R45&Resisc45 &	45	 & N/A	& 31,500&\href{https://www.tensorflow.org/datasets/catalog/resisc45}{Tensorflow}\\

\bottomrule
\end{tabular}
}
\caption{Statistics of datasets used in zero-shot and linear probe. }
\label{table:downstream_ic_dataset}
\end{table*}

\section{Experimental Details}
\textbf{Dataset.}
Table~\ref{table:downstream_ic_dataset} describes the statistics of dataset used for evaluation. We pick the test datasets based on UniCL~\citep{yang2022unified} and availability in \href{https://www.tensorflow.org/datasets/catalog/}{Tensorflow dataset}. 
We use the test set to evaluate zero-shot recognition and linear probe while the train set is used to train a linear classifier. Note that since EuroSAT and Resisc45 utilize the training split for evaluation, we exclude the two datasets from linear probe evaluation. Also, since Oxford Flowers do not have many training samples (10 samples per class), we exclude the dataset from the evaluation too. 

\textbf{Data Augmentation.}
Following UniCL~\citep{yang2022unified}, only random cropping is applied to train all models for a fair comparison.

\textbf{Computation.}
We use 32 Nvidia Tesla V100 GPUs to train all models. 4 nodes, where each node has 8 GPUs, are used to run  experiments.

\section{Additional Results}

\begin{figure*}
    \centering
    \includegraphics[width=\textwidth]{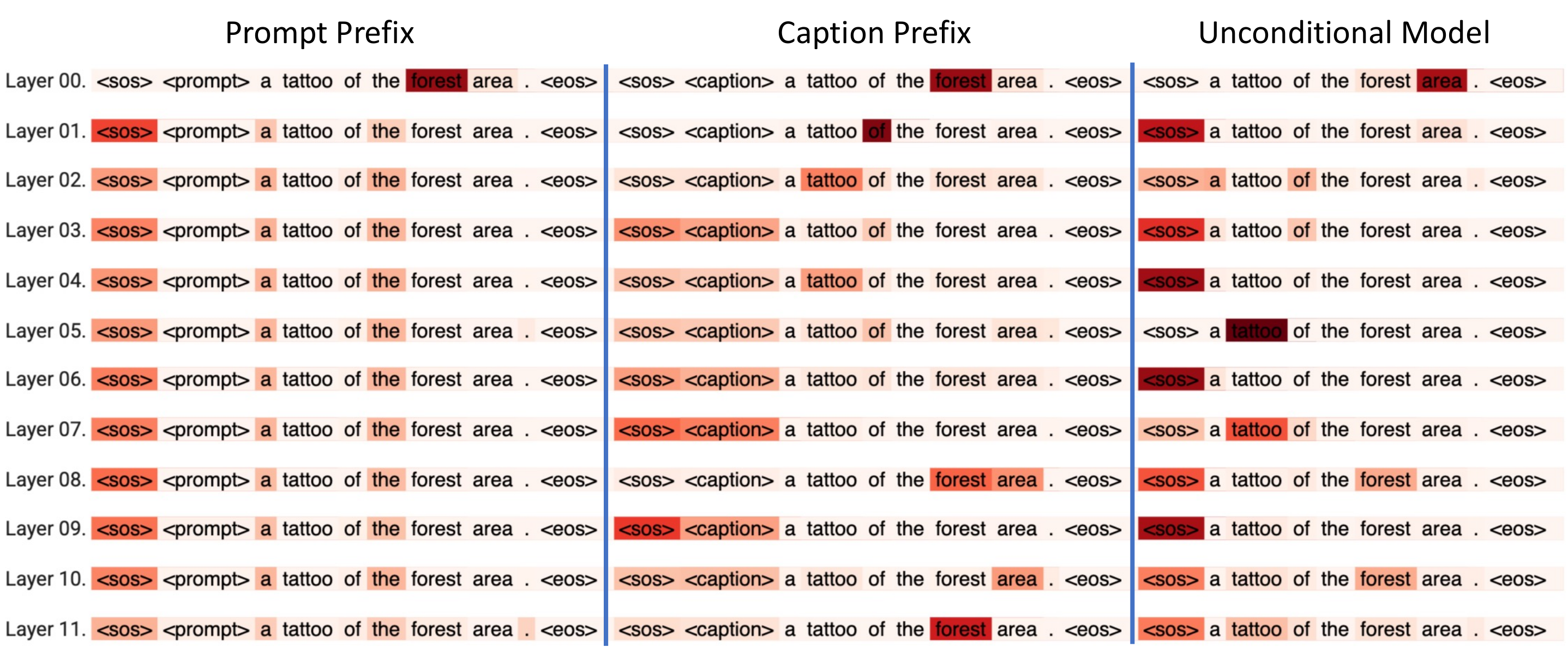}
    \caption{Attention visualization for a \textit{class prompt}. Note that the attention weights are for and end token. Best viewed in color. The class name shown here is one of class prompts in the EUROSAT dataset. Different rows show the weights of different transformer layers. With a prompt prefix (leftmost), the model focuses on a class name (\textit{forest area}) while caption prefix (middle) allows a model to pay attention to another noun, \textit{tattoo}. By prefix conditioning, the attention of the model changes as intended.}
    \label{fig:attention_prompt_appendix}
\end{figure*}
\begin{figure*}
    \centering
    \includegraphics[width=\textwidth]{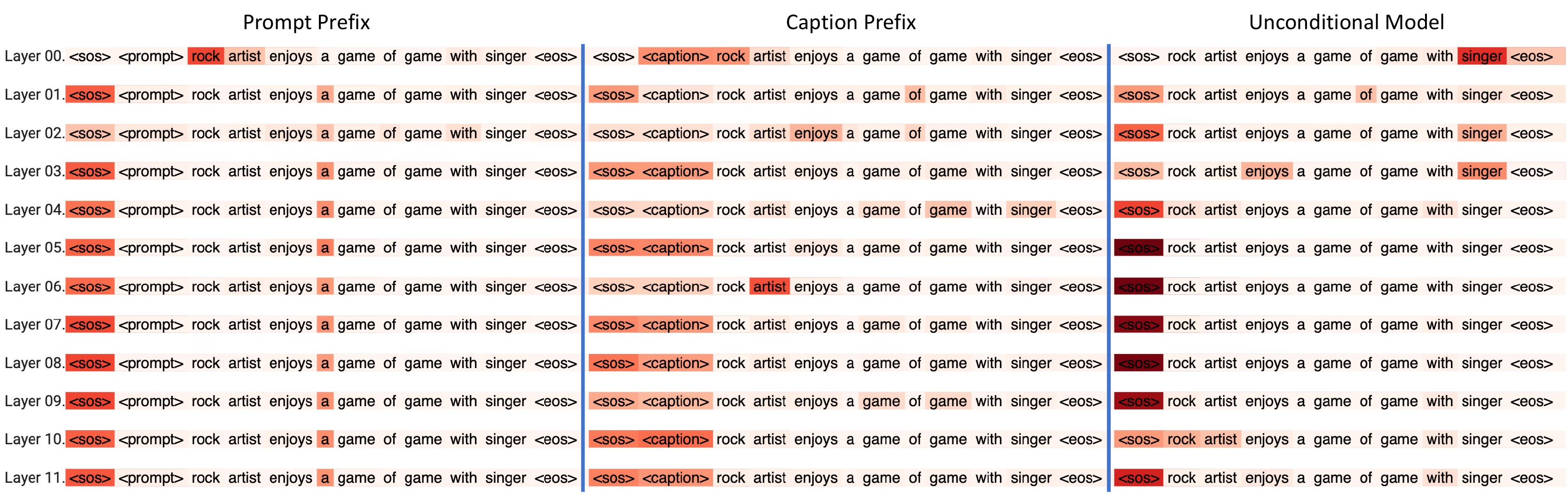}
    \caption{Attention visualization for a \textit{real caption}. Note that the attention weights are for and end token. Best viewed in color. The sentence shown here is from CC3M. Different rows show the weights of different transformer layers. Caption prefix conditioning helps to attend to many words while prompt conditioning fails to do that.}
    \label{fig:attention_caption_appendix}
\end{figure*}
\textbf{Attention Visualization.}
Fig.~\ref{fig:attention_prompt_appendix} visualizes attention weights for the class \textit{forest area}, where a prompt template, \textit{a tatto of}, is employed. The model focuses on a word, \textit{forest} when prompt prefix is employed. In other two cases, the model also pays much attention to \textit{tatoo} probably because the word should provide useful information to distinguish a sentence from others for image-caption contrastive learning. 
Fig.~\ref{fig:attention_caption_appendix} represents attention for a real caption from CC3M. While the model conditioned with caption prefix and unconditional model attend to several words through many layers, the model conditioned with prompt prefix shows clear attention only in the first layer. Since the prompt-conditioned model has never seen the real caption during training, it fails in attending to discriminative words.

\begin{table*}[ht]
    \centering
    \footnotesize
    \begin{tabular}{c|c|c|c|cc|cc}
    \toprule
    \multirow{2}{*}{Train Data}&\multirow{2}{*}{\makecell{Train on \\ Synonym}}& \multirow{2}{*}{\makecell{Prefix\\ Training}}&\multirow{2}{*}{\makecell{Test-Time \\ Prefix}}&\multicolumn{2}{c|}{Original}&\multicolumn{2}{c}{Synonym}\\
    &&&&top-1&top-5&top-1&top-5\\
     \midrule
    IN1K + CC12M &&&N/A&69.3&89.3&31.2&49.5\\
    IN1K + CC12M&&\checkmark&Prompt&\bf{75.0}&\bf{92.9}&\bf{38.3}&54.8\\
    IN1K + CC12M&&\checkmark&Caption&71.4&91.6&36.6&\bf{56.7}\\
    \midrule
    IN21K + CC12M &&&N/A&54.5&83.2&23.1&43.9\\ 
    IN21K + CC12M &&\checkmark&Prompt&\bf{69.9}&\bf{92.4} &32.1&53.7\\
    IN21K + CC12M &&\checkmark&Caption& 65.3&90.6 &\bf{33.5}&\bf{56.9}\\
     \midrule
      IN21K + CC12M &\checkmark&\checkmark&Prompt&54.4&78.6&\bf{70.8}&\bf{92.8}\\
     IN21K + CC12M &\checkmark&\checkmark&Caption& \bf{54.5}&\bf{82.6} &59.0&86.1\\
        \bottomrule
    \end{tabular}
    \caption{Evaluation on the robustness to the class name shift using ImageNet-1K. \textit{Original} refers to the subset of ImageNet-1K classes while \textit{synonym} refers to their synonyms taken from Wordnet. The last two rows indicate the models trained with the synonyms, thus showing superior performance on \textit{synonym} whereas degrading performance on \textit{Original}.}
    \label{tab:vocabulary_shift}
\end{table*}

\begin{table*}[ht]
 \centering
\setlength{\tabcolsep}{2.5pt}
\begin{tabular}{c|c|cccc|cccc}
 \toprule
\multirow{2}{*}{\makecell{Prefix \\Training}} &\multirow{2}{*}{\makecell{Test-time \\ Prefix}}&\multicolumn{4}{c|}{CC3M}      & \multicolumn{4}{c}{COCO}      \\
&&I2T@1 & I2T@5 & T2I@1 & T2I@5 & I2T@1 & I2T@5 & T2I@1 & T2I@5 \\\hline
&N/A&21.8 &47.4 & 21.0 &45.7   & 23.9  & 49.5  & 18.7  & 43.2  \\
\checkmark&Prompt&13.1&31.3&8.1&21.8&17.2&38.1&16.8&37.7\\
\checkmark&Caption&\bf{22.6}  & \bf{47.5}  & \bf{21.6}  & \bf{46.1}  & \bf{24.7}  & \bf{49.7}  & \bf{19.7}  & \bf{43.9} \\
 \bottomrule
\end{tabular}
\vspace{2mm}
\caption{Image-text retrieval results on CC3M and COCO. The performance is evaluated on the subset of CC3M and validation set of COCO. All models are trained on CC12M and ImageNet-21K. Caption conditioning (last row) slightly improves retrieval performance compared to the unconditional model (first row). Since prompt conditioning (middle) tailors a model for class-prompt, it fails to extract discriminative information from real captions.}
\label{tab:image_text_retrieval}
\end{table*}

\textbf{Class Name Shift.}
Test samples can be unseen with respect to image classification data in two ways (or combinations of two): 1) The image is similar to training distribution, but the class name used for testing is different from the image classification label. 2) Although the class label is the same, the image data comes from the different distributions. Datasets evaluated in the zeros-shot recognition include both two cases since class names and images are from different domains. 2) is analyzed in Subsection~4.3 of the main paper, \textit{Robustness in image domain shift}. We analyze 1) by evaluating the recognition performance of ImageNet-1K by changing its class name from the one used during training. We find a synonym for each class with WordNet~\citep{miller1995wordnet}, where we exclude synonyms substantially similar to the original class name and obtain synonyms for 525 classes. Then, we use the synonym to classify images during evaluation. Since the input image distribution does not vary, we can evaluate the performance on the class name shift. If the model is robust to the change in the class name, the degrade in the performance should be small. 

The first 6 rows of Table~\ref{tab:vocabulary_shift} describe the models trained with the original class names and evaluated on both original ones and synonyms, and the last two rows represent a model trained with synonyms, where the original class names are replaced with synonyms.
Prompt prefix outperforms caption prefix with a large margin in testing with class names used in training time. Generally, caption prefix performs better when tested with the class names different from the ones used during training. Prompt prefix is tailored to handle class names employed during training time while caption prefix enables the language encoder to extract more general representations. 

Interestingly, the choice of class names seems to significantly change the generalization as shown in the comparison between a model trained with synonyms and original class names. The original model decreases the accuracy more than 30\% by changing the class name while the model trained with synonym decreases less than 20\%.

\textbf{Image-Caption Retrieval.}
In Table~\ref{tab:image_text_retrieval}, we evaluate the performance of image-caption retrieval using the subset of CC3M (12288 pairs of image and caption) and COCO validation set (5000 pairs of image and caption), where all models are trained with CC12M and ImageNet-21K. First, our model (last row) slightly performs better than the model without conditioning (first row). Second, prompt prefix conditioning (second row) significantly performs worse than caption prefix conditioning (last row). Since the prompt prefix conditioning specializes a model for the class name prompts of ImageNet21K, the conditioning does not generalize well to real captions.  

\textbf{Larger Batch-size and Training Epochs.}
We examine the effect of increasing batch-size and training epochs in Table~\ref{tab:epochs_bsize}. In CLIP, increasing the batch-size and training epochs improves the performance of both ImageNet-1K and zero-shot recognition. On the other hand, the zero-shot performance of UniCL is not benefited from training with longer epochs (compare last and second to last row). UniCL attempts to ensure the invariance of images from the same classes by supervised contrastive loss while CLIP does not consider it. However, such invariance is not necessarily required in zero-shot recognition, which leads to the degraded performance.

\textbf{Comparison to Reported UniCL's Results.}
In the main paper, we provide our reproduced results of UniCL, which is based on our implementation, since the authors have not released the code and did not report the numerical accuracy of each zero-shot recognition. 
In this paragraph, we compare our approach and the reported performance of UniCL~\citep{yang2022unified} and K-Lite~\citep{shen2022k} by aligning several hyper-parameters, e.g., batch-size and training epochs, using ImageNet-1K. When using ImageNet-22K and CC-15M for training, our method (batch-size:4096, training epochs: 30) shows 73.9 while UniCL (batch-size:4096, training epochs 32) reports 71.5. When using ImageNet-21K excluding ImageNet-1K and CC-15M, our method (batch-size:1024, training epochs 30) shows 49.7 whereas UniCL (batch-size: 4096, training epochs: 32) and K-Lite (batch-size: 4096, training epochs: 32) perform 46.6 and 48.7 respectively according to K-Lite results (See last two rows of Table 3 in~\citep{shen2022k}). These results suggest that our method performs better than the reported numbers of UniCL and K-Lite in ImageNet-1K. Also, the knowledge augmentation technique proposed by K-Lite can be complementary to our approach, thus combining two approaches is an interesting research direction.

\begin{figure}
     \centering
     \begin{subfigure}[b]{0.44\textwidth}
         \centering
         \includegraphics[width=\textwidth]{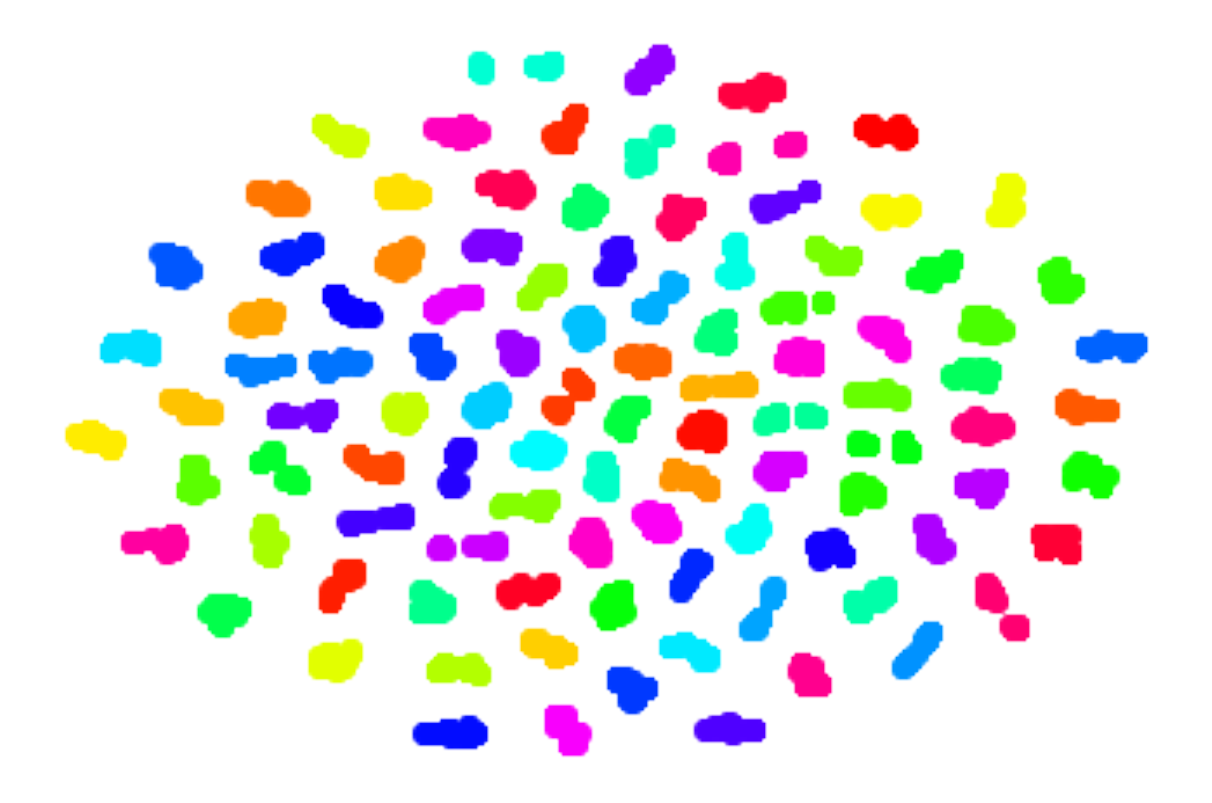}
         \caption{Prompt conditioned}
         \label{fig:in1k_prompt}
     \end{subfigure}
     \begin{subfigure}[b]{0.44\textwidth}
         \centering
         \includegraphics[width=\textwidth]{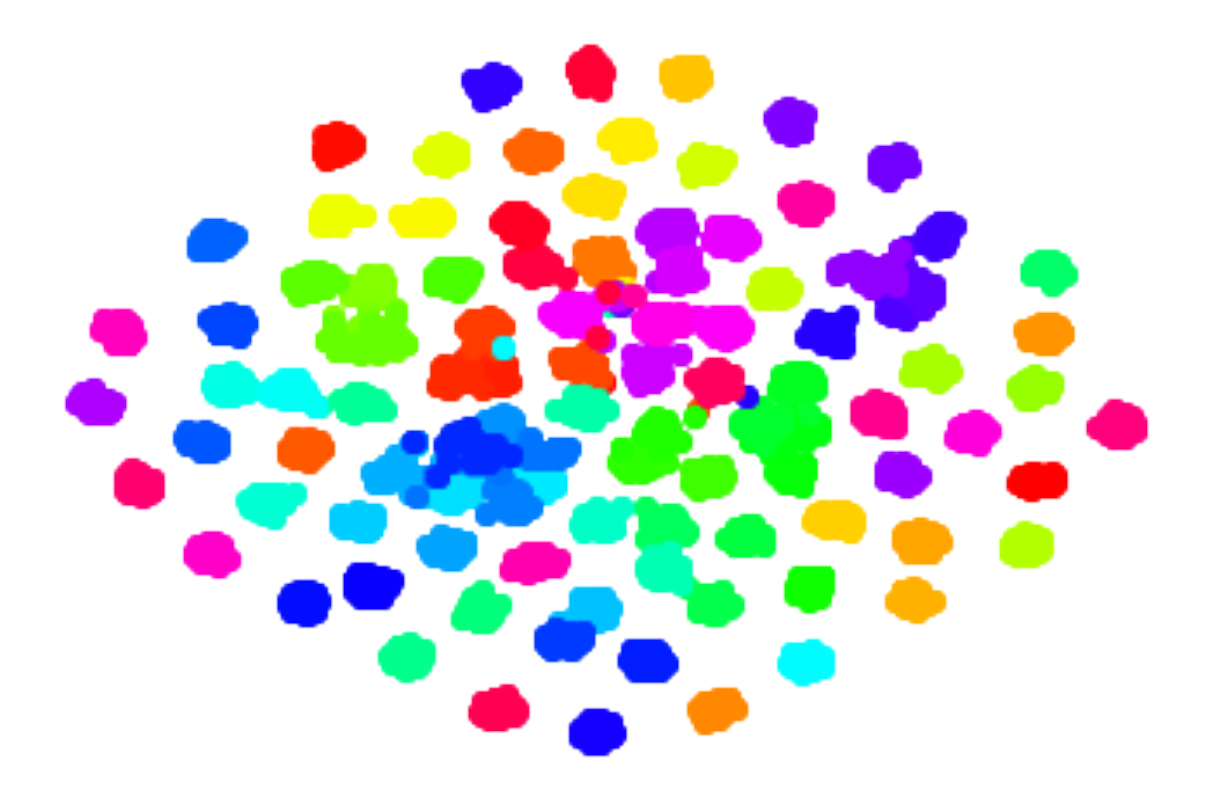}
         \caption{Caption conditioned}
         \label{fig:in1k_caption}
     \end{subfigure}
      \caption{T-SNE~\cite{van2008visualizing} visualization of the class-prompt features of ImageNet-1K with different prefix conditions. Different colors indicate language embeddings of different classes. Prompt conditioning extracts more class discriminative representations than caption conditioning.}
      \label{fig:tsne_in1k}
\end{figure}

\textbf{T-SNE visualization for language features.}
Fig.~\ref{fig:tsne_in1k} visualizes extracted language features (ImageNet-1K) conditioned with different prefixes. 
The prompt-prefix (left) has lower intra-class and higher inter-class variance, whereas caption-prefix (right) shows higher intra-class variance across prompts.

\begin{figure}
     \centering
     \begin{subfigure}[b]{0.44\textwidth}
         \centering
         \includegraphics[width=\textwidth]{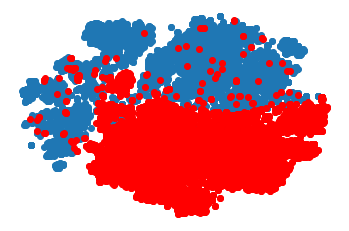}
         \caption{W/O conditioning}
         \label{fig:in1k_vs_cc3m_baseline}
     \end{subfigure}
     \begin{subfigure}[b]{0.44\textwidth}
         \centering
         \includegraphics[width=\textwidth]{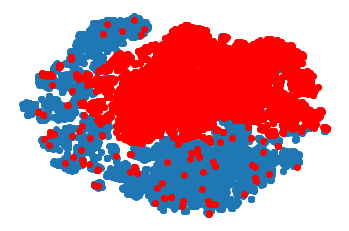}
         \caption{With conditioning}
         \label{fig:in1k_vs_cc3m_condition}
     \end{subfigure}
      \caption{T-SNE~\cite{van2008visualizing} visualization of the image features of ImageNet-1K (blue) and CC3M (red). Since ImageNet-1K is object-centered while CC3M covers more diverse scenes, the distributions are separated. This is consistent across baseline (w/o conditioning) and our method (with conditioning).}
      \label{fig:tsne_in1k_vs_cc3m}
\end{figure}

\textbf{T-SNE visualization for image features.}
Fig.~\ref{fig:tsne_in1k_vs_cc3m} visualizes image features from ImageNet-1K (blue) and CC3M (red). Since ImageNet-1K is object-centered while CC3M covers more diverse scenes, the distributions are separated. This is consistent across baseline (w/o conditioning) and our method (with conditioning).

\begin{figure}
     \centering
     \begin{subfigure}[b]{0.44\textwidth}
         \centering
         \includegraphics[width=\textwidth]{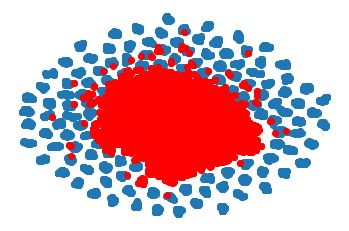}
         \caption{Unconditioned model}
         \label{fig:unconditioned_lang}
     \end{subfigure}
     \begin{subfigure}[b]{0.44\textwidth}
         \centering
         \includegraphics[width=\textwidth]{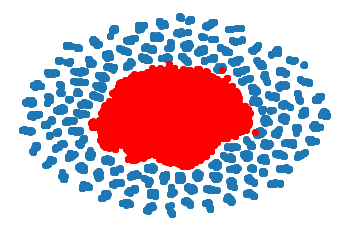}
         \caption{Conditioned model}
         \label{fig:conditioned_lang}
     \end{subfigure}
      \caption{T-SNE~\cite{van2008visualizing} visualization of language features of ImageNet-1K class prompts (Blue) and CC3M captions (Red) for unconditioned (left) and conditioned (right) respectively. Our proposed condition better differentiates prompts from real captions.}
      \label{fig:language_embed_cond_vs_uncond}
\end{figure}

\textbf{Comparison between unconditioned and conditioned model by language features.} Fig.~\ref{fig:language_embed_cond_vs_uncond} visualizes language features of ImageNet-1K class prompts (Blue) and CC3M captions (Red) for unconditioned (left) and conditioned (right) respectively. Note that the conditioned model utilizes prompt prefix for class prompts and caption prefix for real captions respectively.
As seen from the visualization, unconditioned model cannot distinguish some prompts from captions of CC3M. This is probably because some captions are similar to class prompts of ImageNet. By contrast, the conditioned model differentiate class prompts from captions better than unconditioned model due to the prefix conditioning.

\begin{table*}[t]
    \centering
    \resizebox{0.8\linewidth}{!}{%
    \begin{tabular}{ll|c|c|c|cc}
    \toprule
     \multicolumn{2}{c|}{\multirow{2}{*}{Training Data}} &  \multirow{3}{*}{Objective} & \multirow{3}{*}{Batch-size} &\multirow{3}{*}{Epochs}&  \multicolumn{2}{c}{Metric}  \\
        \cmidrule{6-7}
         \multirow{2}{*}{Classification} & \multirow{2}{*}{Caption} &&&& \multirow{2}{*}{IN-1K} & \multirow{2}{*}{\makecell{Zero-shot \\ 11 datasets}}  \\
        & &  &&&\\
        \midrule
        ImageNet-21K & CC-12M  &  CLIP &1024&15&67.3&57.8\\ 
        ImageNet-21K & CC-12M  &  CLIP &1024&30&\bf{69.1}&\bf{58.3}\\ 
         \midrule
        ImageNet-22K & CC-15M  &  CLIP & 1024&15& 69.3 & 58.5\\
        ImageNet-22K & CC-15M  &  CLIP & 4096&15& 71.1 & 59.5\\
        ImageNet-22K & CC-15M  &  CLIP & 4096 & 30 &\bf{72.2} & \bf{59.8}\\
        \midrule
        ImageNet-22K & CC-15M  &  UniCL  &1024&15& 69.7 & 58.5\\
        ImageNet-22K & CC-15M  &  UniCL  &4096&15& 70.3 & \bf{60.4}\\
        ImageNet-22K & CC-15M  &  UniCL  &4096&30& \bf{73.9} & 58.9\\
        \bottomrule
    \end{tabular}
    }
    \caption{Performance comparison among different batch-size and training epochs. ImageNet-22K denotes the combination of ImageNet-21K and ImageNet-1K, CC-15M indicates that of CC-12M and CC-3M.}
        \label{tab:epochs_bsize}
\end{table*}


{\small
\bibliographystyle{ieee_fullname}
\bibliography{reference}
}

\end{document}